\pdfoutput=1

\documentclass[11pt]{article}

\usepackage[]{acl}

\usepackage{times}
\usepackage{latexsym}

\usepackage[T1]{fontenc}

\usepackage[utf8]{inputenc}

\usepackage{microtype}

\usepackage{inconsolata}
\usepackage{subfigure}
\usepackage{graphicx}
\usepackage{amsmath}
\usepackage{amssymb}
\usepackage{amsthm}
\usepackage{multirow}
\usepackage{multicol}

\usepackage[ruled,linesnumbered]{algorithm2e}
\usepackage[perpage]{footmisc}
\usepackage{float}
\usepackage{enumitem}
\usepackage[section]{placeins}

\title{EMS-SD: Efficient Multi-sample Speculative Decoding \\ for Accelerating Large Language Models}

\author{
    \begin{minipage}{\linewidth}
    \begin{center}
    \large 
    Yunsheng Ni \hspace{4pt} 
    Chuanjian Liu \hspace{4pt}
    Yehui Tang \hspace{4pt}
    Kai Han$^*$ \hspace{4pt}  
    Yunhe Wang\Thanks{Corresponding author} \\
    \scalebox{0.75}{Huawei Noah’s Ark Lab}\\
    \scalebox{0.75}{\texttt{\{niyunsheng,kai.han,yunhe.wang\}@huawei.com}}
    \end{center}
    \end{minipage}
 }

\begin{document}
\maketitle
\begin{abstract}
Speculative decoding emerges as a pivotal technique for enhancing the inference speed of Large Language Models (LLMs).
Despite recent research aiming to improve prediction efficiency, multi-sample speculative decoding has been overlooked due to varying numbers of accepted tokens within a batch in the verification phase.
    Vanilla method adds padding tokens in order to ensure that the number of new tokens remains consistent across samples.
    However, this increases the computational and memory access overhead, thereby reducing the speedup ratio.
    We propose a novel method that can resolve the issue of inconsistent tokens accepted by different samples 
    without necessitating an increase in memory or computing overhead.
    Furthermore, our proposed method can handle the situation where the prediction tokens of different samples are inconsistent without the need to add padding tokens.
    Sufficient experiments demonstrate the efficacy of our method.
    Our code is available at \href{https://github.com/niyunsheng/EMS-SD}{https://github.com/niyunsheng/EMS-SD}.
\end{abstract}

\section{Introduction}

Large Language Models (LLMs) ~\cite{radford2019language, achiam2023gpt, touvron2023llama, wang2023pangu} have demonstrated considerable capabilities, particularly in the realm of natural language processing.
Autoregressive Large Language Models generate a token in a single pass, 
whereas speculative decoding allows large models to generate multiple tokens in a single pass, 
thereby greatly improving inference speed.
It is crucial to highlight that the inference time of LLMs on a single token and multiple tokens is approximate.
Consequently, reducing the number of inference steps can significantly reduce the inference time.

A plethora of efficient speculative decoding methods have been proposed recently.
However, none of these methods provide a comprehensive study of speculative decoding in multi-sample scenarios. To the best of our knowledge, only EAGLE~\cite{li2024eagle} presents results for batch sizes $\le 4$ but doesn't discuss larger batch sizes.

The primary challenge in multi-sample speculative decoding is the inconsistency in the number of accepted tokens across samples following a single inference. 
Vanilla solution is to add padding tokens in order to achieve uniformity. 
This approach is also employed by EAGLE.
Nevertheless, these padding tokens increase the computational and memory access overhead, which becomes significant as batch size increases, thereby reducing speedup ratio.

\begin{figure}[t]
    \centering
	\includegraphics[width=0.98\linewidth]{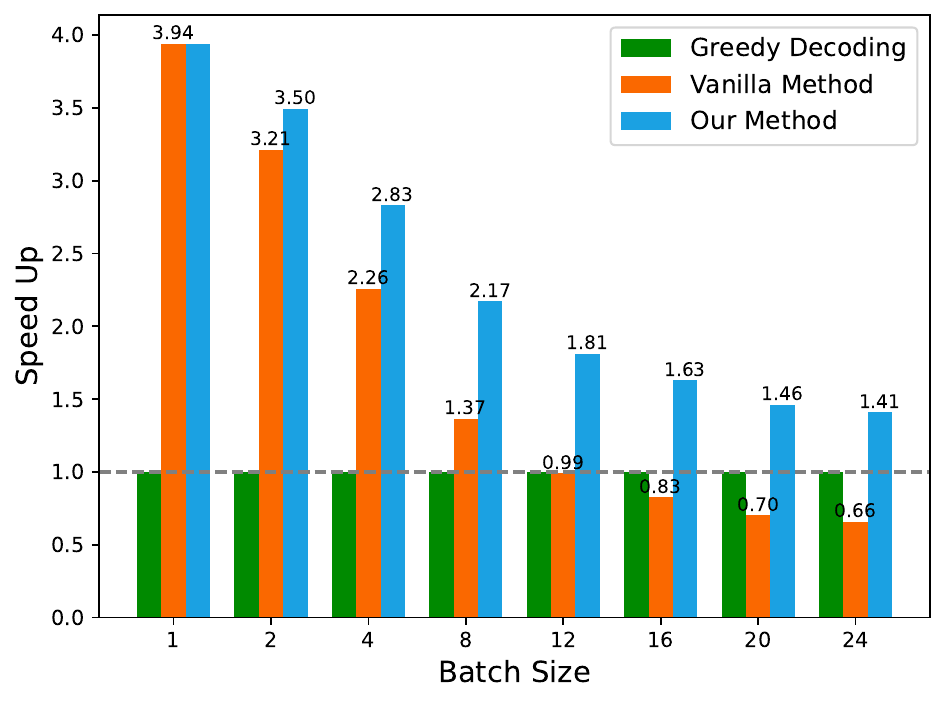}
	\vspace{-8pt}
    \caption{
    Speedup ratio of Opt-6.7b on the CNN/Daily Mail Dataset for greedy settings when batch size $\ge 1$, 
    utilizing LLMA~\cite{yang2023inference} as the basic speculative decoding method. 
    Our method demonstrates superior performance to the vanilla method under varying batch sizes. 
    The larger the batch size, the more pronounced the advantage of our method.}
    \label{img:fig1_vanilla_vs_ours}
    \vspace{-15pt}
\end{figure}

\begin{quote}
    \emph{Can we perform multi-sample speculative decoding without increasing computational and memory access overhead?}
\end{quote}

We proposed a novel and efficient method to resolve this issue.
Specifically, we proposed unpad Key-Value (KV) cache in the verification phase, which specifies the start locations of the KV cache for different samples, thus eliminating the need for padding tokens.
Furthermore, in anticipation of the potential discrepancy in the number of predicted tokens across different samples, we proposed the unpad input tokens method as a solution in the prediction phase.
This method concatenates all input tokens prior to inference and expands these tokens during the calculation of attention.

The main contributions are as follows:
\begin{enumerate}[itemsep=0pt,topsep=0pt,parsep=0pt]
    \item We proposed an Efficient Multi-sample Speculative Decoding method (EMS-SD), which takes full account of the inhomogeneity between different samples.
    Even if the new generated token numbers of different samples vary, the KV cache is continuous without the addition of padding tokens.
    Similarly, when the prediction token numbers of different samples vary, all input tokens are spliced without the addition of padding tokens.
    \item Sufficient experiments have proven that our proposed method achieves a much higher speedup than vanilla methods in multi-sample speculative decoding.
    \item We are the first to study speculative decoding in the context of multi-sample situations, and we have proposed an effective method for addressing this issue.
    Our method can be easily integrated into almost all basic speculative decoding methods.
\end{enumerate}

\section{Related Works}

\textbf{Large Language Models.}
Since the advent of the GPT~\cite{radford2019language} series of models, particularly after the emergence of ChatGPT~\cite{achiam2023gpt}, there has been a proliferation of large language models, including Llama~\cite{touvron2023llama}, Vicuna~\cite{chiang2023vicuna}, ChatGLM~\cite{zeng2022glm}, QWen~\cite{bai2023qwen}, Baichuan~\cite{yang2023baichuan}, Gemini~\cite{team2023gemini}, Pangu-$\pi$~\cite{wang2023pangu}, Mistral~\cite{jiang2023mistral, jiang2024mixtral}, etc. 

\noindent \textbf{Speculative decoding.}
Speculative decoding can be divided into two stages in general: prediction and verification.
Some studies have proposed efficient prediction methods.
These prediction methods can be broadly classified into two categories: 
those that require training and those that do not.
For example, methods that do not require training include LLMA~\cite{yang2023inference}, REST~\cite{he2023rest}, Lookahead~\cite{fu2023lookahead}, PLD~\cite{saxena2023prompt}, etc.
In contrast, methods that require training include draft model prediction~\cite{leviathan2023fast}, Medusa~\cite{cai2024medusa}, Hydra~\cite{ankner2024hydra}, kangaroo~\cite{liu2024kangaroo}, EAGLE~\cite{li2024eagle}, etc.

\noindent \textbf{Dynamic Tree decoding.}
SpecInfer~\cite{miao2023specinfer} 
introduces tree decoding mechanism, which predicts multiple tokens at the same position to improve the acceptance rate.
The tree structure is manually designed, and so is Medusa, EAGLE, etc. 
Some recent studies have focused on the problem of dynamic tree decoding.
Sequoia~\cite{chen2024sequoia} introduces a hardware-aware tree optimizer.
RSD~\cite{jeon2024recursive} dynamically modifies the tree structure within fixed computational budgets.
And EAGLE2~\cite{li2024eagle} generates the prediction tree dynamically based on confidence scores from the draft model.

\section{Approach}

\begin{figure*}[ht]
    \centering
    \includegraphics[width=0.95\linewidth]{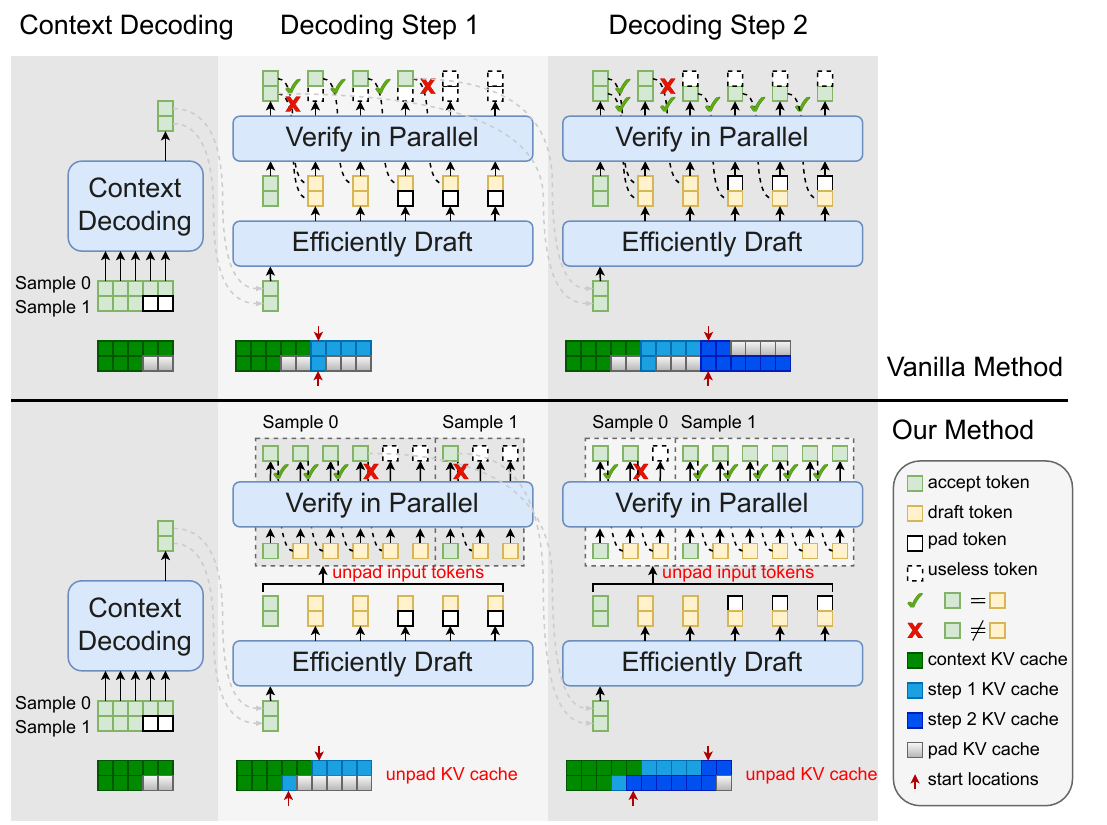}
    \caption{
    Our Method v.s. Vanilla Method. 
    We specify the location of the KV cache for each sample individually, thus eliminating the necessity for the addition of padding to the KV cache.
    And we concatenate all input tokens of each sample into a single sequence without padding tokens when the number of prediction tokens differs between samples.
    Our method demonstrates superior performance than the vanilla method, without the need for additional computational and memory access overhead.
    }
    \label{fig:method}
    \vspace{-15pt}
\end{figure*}

\begin{table*}[ht]
  \caption{The two constructed samples demonstrate in Figure~\ref{fig:method}.
  During the two decoding steps, the number of tokens predicted and accepted by the two samples differs.
  If the vanilla method is employed, it's necessary to incorporate padding tokens in both predication and verification phases of speculative decoding.
  }
  \label{tab:case_analysis}
  \centering
\begin{tabular}{c|cc|cc|cc|cc}
    \hline
    \multirow{3}{*}{\textbf{Sample}} & \multicolumn{4}{c}{\textbf{Decoding Step 1}} & \multicolumn{4}{|c}{\textbf{Decoding Step 2}} \\
    & \multicolumn{2}{c}{\textbf{Predication Phase}} & \multicolumn{2}{|c}{\textbf{Verification Phase}} & \multicolumn{2}{|c}{\textbf{Predication Phase}} & \multicolumn{2}{|c}{\textbf{Verification Phase}} \\
    & Predict & Padding & Accept & Padding & Predict & Padding & Accept & Padding \\
    \hline
    0 & 5 & 0 & 4 & 0 & 2 & 3 & 2 & 4 \\ 
    1 & 2 & 3 & 1 & 3 & 5 & 0 & 6 & 0 \\
    \hline
  \end{tabular}
\end{table*}

\subsection{Rethinking Vanilla Multi-sample Speculative Decoding}
\label{sec:vanilla_method}

\noindent \textbf{Restrictions on memory access.}
It should be noted that mainstream AI frameworks such as PyTorch~\cite{paszke2019pytorch} only support \textbf{aligned key-value cache access}.
Consequently, two key requirements must be met for LLMs inference:
(1) the number of tokens across different samples with in a batch must be equal prior to inference, and (2) the input token count must remain consistent for all samples during inference.
To ensure uniformity, padding tokens are added to samples with varying token lengths. Additionally, attention masks are used to prevent the computation of padding tokens.

\noindent \textbf{Add padding tokens to align the output lengths of different samples.}
The primary issue is that the number of accept tokens during the verification phase varies considerably between samples within a batch.
To illustrate, if $k$ tokens are predicted in the prediction stage, then the number of accept tokens can be varied from $1$ to $k+1$.
Vanilla method adds padding tokens to ensure that the number of new tokens is the same for each sample within in a batch.
Nevertheless, this approach leads to a considerable increase in the computational and memory access overhead, 
which in turn results in a significant reduction in the speedup.
In Appendix~\ref{sec:appendix_analysis_vanilla_padding_tokens}, 
we present a theoretical analysis of the impact of padding tokens on speedup.

\noindent \textbf{Add padding tokens to align the input lengths of different samples.}
A further issue is that the number of predicted tokens for different samples in the prediction stage may vary.
In this case, padding token also needs to be added to align the input lengths.
This issue does not arise in all circumstances, and is most commonly observed in retrieval-based prediction scenarios, including LLMA~\cite{yang2023inference} and REST~\cite{he2023rest}.
This is due to the fact that the retrieval-based prediction method employs a text matching process, whereby different samples may not be able to match the predicted text simultaneously.
In more general methods, such as draft model prediction~\cite{leviathan2023fast}, 
generate same number of prediction tokens for different samples.
Some recent studies have focused on the problem of dynamic tree decoding~\cite{chen2024sequoia,jeon2024recursive}.
It is possible that in the future, there may be different optimal prediction trees or optimal numbers of tokens for different samples.

\noindent \textbf{Case analysis.}
As illustrated in Figure~\ref{fig:method} and Table~\ref{tab:case_analysis}, 
we construct two samples within a batch as an example.
In the decoding step 1, sample 1 have to add 3 padding tokens in order to ensure that the input lengths are identical to those of sample 0.
Subsequently, following the verification phase, sample 1 must add 3 padding tokens in the KV cache in order to ensure that the output lengths are identical to those of sample 0.
In the decoding step 2, sample 0 have to add 3 padding tokens during the predication phase and 4 padding tokens in the KV cache subsequent to the verification phase.

\subsection{Efficient Multi-sample Speculative Decoding}

Vanilla Method tends to result in elevated computational and memory access overheads. 
In contrast, our approach does not entail such drawbacks, thereby conferring a higher speedup ratio.
In this section, we first point out that aligned KV cache access is not immutable, and then present two key components of our approach: unpad KV cache and unpad input tokens.

\noindent \textbf{Aligned KV cache access is not mandatory.}
In autoregressive models, each token is conditioned only on preceding tokens during the attention computation.
Theoretically, given the location of the input token and access to the KV cache, we can calculate the attention output.
These operations can be encapsulated within CUDA kernels, as evidenced by implementations in frameworks such as FasterTransformer~\cite{NVIDIA2021FasterTransformer}, FlashAttention~\cite{dao2022flashattention}, and PyTorch~\cite{paszke2019pytorch}~\footnote{These frameworks provide the basic CUDA kernel for computing attention output. We need to modify these kernels to implement our method for supporting speculative decoding in multi-sample situations.}.
When invoking these kernels, we can compute attention output for different tokens, even if these different tokens are in different samples and rely on different numbers of preceding tokens.

\noindent \textbf{Unpad KV cache.}
Firstly, we introduce the first major component: unpad KV cache. 
This eliminates the need to add padding tokens when different samples accept different lengths in the verification phase.
In particular, we specify the start location of the KV cache for each sample individually, rather than aligning writes in a manner similar to Pytorch.
It should be noted that the varying start locations of samples lead to slight discrepancies in the computational workload for the attention CUDA kernels.
Nevertheless, since all tokens across varying positions and samples compute their attention outputs in parallel, the overall speed is dictated by the token necessitating the greatest computational load, typically the one with the highest number of preceding tokens.
As illustrated in the lower part of Figure~\ref{fig:method}, 
the start locations of KV cache of the two samples is distinct. 
For each input token, we initially compute its KV cache and subsequently write it to memory based on the specified position for each sample. Thereafter, the attention outputs for all tokens, across various samples and positions, are calculated in parallel.

By employing unique KV cache start positions for each sample, we can independently determine the subsequent start location during verification, regardless of varying acceptance lengths across samples.
Consequently, this approach negates the need for extra padding tokens, thereby preventing memory waste and computational overhead.
As shown in Figure~\ref{fig:method}, sample 0 accepted 4 tokens, advancing the KV cache start location by 4. While sample 1 accepted 1 token, advancing it by 1.

\begin{figure}[h]
    \centering
    \includegraphics[width=0.9\linewidth]{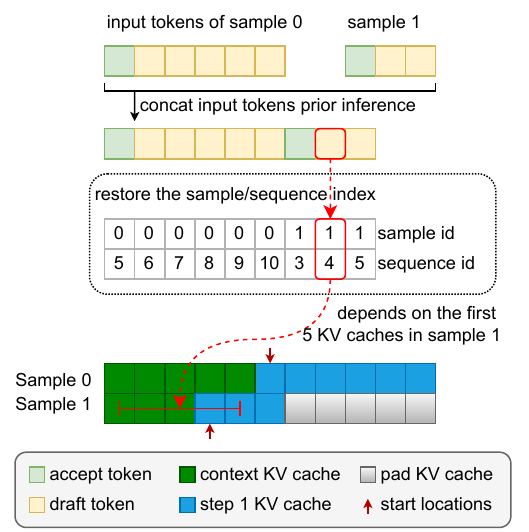}
    \caption{
    The detailed processing of unpad input tokens of decoding step 1 in Figure~\ref{fig:method}.
    Sample 0 predicted 5 tokens, while sample 1 predicted 2 tokens. All tokens are concatenated before inference, and the sample/sequence index is restored when attention is computed within the CUDA kernels.
    Consequently, each token is aware of the specific KV caches to which it can utilize for parallel computation.
    }
    \label{fig:unpad_input_tokens} 
    \vspace{-15pt}
\end{figure}

\noindent \textbf{Unpad input tokens.}
Secondly, in order to address the issue of differing numbers of input tokens across different samples, we proposed the "unpad input tokens" method as a solution.
In general, prior to inputting into the Transformer network, all input tokens are concatenated together, and the number of input tokens for each sample is recorded. 
Additionally, during the attention result calculations, the CUDA kernel reconstructs the original batch indices and sequence positions for each token. This reconstruction enables us to identify the specific KV cache that each token needs to rely on.
Figure~\ref{fig:unpad_input_tokens} shows the general processing flow.
Refer to Appendix~\ref{sec:appendix_algorihtm_of_unpad_input_tokens} for specific processing procedures.

\begin{table*}[ht]
    \centering
    \caption{
    Ablation study using LLMA were conducted on two key methods: unpad KV cache and unpad input tokens. Under different batch sizes, our method demonstrated a significantly higher acceleration ratio than vanilla method. The method "unpad KV cache" played a more prominent role.TPS stands for tokens per second.
    }
    \begin{tabular}{cccccc}
        \hline
        \multirow{2}{*}{\textbf{Batch Size}} & \multirow{2}{*}{\textbf{Inference Method}} & \textbf{Unpad} & \textbf{Unpad} & \multirow{2}{*}{\textbf{TPS}} & \multirow{2}{*}{\textbf{Speed up}} \\ 
        ~ & ~ & \textbf{KV Cache} & \textbf{Input Tokens} & ~ & ~ \\
        \hline
        \multirow{5}{*}{2} & Greedy Decoding &  &  & 137.49  & ~ \\
        ~ & Vanilla Method &  &  & 441.64  & 3.21  \\
         \cline{2-6} 
         & \multirow{3}{*}{Our Method} & $\checkmark$ &  & 439.40  & 3.20  \\
        ~ & ~ &  & $\checkmark$ & 477.72  & 3.47  \\
        ~ & ~ & $\checkmark$ & $\checkmark$ & 480.59  & \textbf{3.50}  \\
        \hline
        \multirow{5}{*}{4} & Greedy Decoding &  &  & 257.37  & ~ \\
        ~ & Vanilla Method &  &  & 581.54  & 2.26  \\
         \cline{2-6} 
         & \multirow{3}{*}{Our Method} & $\checkmark$ &  & 610.41  & 2.37  \\
        ~ & ~ &  & $\checkmark$ & 728.86  & 2.83  \\
        ~ & ~ & $\checkmark$ & $\checkmark$ & 729.17  & \textbf{2.83}  \\ 
        \hline
        \multirow{5}{*}{8} & Greedy Decoding &  &  & 468.89  & ~ \\
        ~ & Vanilla Method &  &  & 640.58  & 1.37  \\
         \cline{2-6} 
         & \multirow{3}{*}{Our Method} & $\checkmark$ &  & 687.11  & 1.47  \\
        ~ & ~ &  & $\checkmark$ & 948.71  & 2.02  \\
        ~ & ~ & $\checkmark$ & $\checkmark$ & 1017.75  & \textbf{2.17}  \\
        \hline
        \multirow{5}{*}{16} & Greedy Decoding &  &  & 774.59  & ~ \\
        ~ & Vanilla Method &  &  & 640.94  & 0.83  \\
         \cline{2-6} 
         & \multirow{3}{*}{Our Method} & $\checkmark$ &  & 734.86  & 0.95  \\
        ~ & ~ &  & $\checkmark$ & 1134.25  & 1.46  \\
        ~ & ~ & $\checkmark$ & $\checkmark$ & 1264.07  & \textbf{1.63}  \\
        \hline
        \multirow{5}{*}{24} & Greedy Decoding &  &   & 936.45  & ~ \\
        ~ & Vanilla Method &  &  & 616.19  & 0.66  \\
         \cline{2-6} 
         & \multirow{3}{*}{Our Method} & $\checkmark$ &  & 708.53  & 0.76  \\
        ~ & ~ &  & $\checkmark$ & 1150.16  & 1.23  \\
        ~ & ~ & $\checkmark$ & $\checkmark$ & 1321.45  & \textbf{1.41} \\
        \hline
    \end{tabular}
    \label{tab:llma_ablation}
    \vspace{-15pt}
\end{table*}

\section{Experiments}

\begin{table*}
    \centering
    \caption{The efficacy of our method evaluated on two smaller models using LLMA. Our method demonstrates superior performance on different batch sizes and models of varying sizes. When the batch size increases, the speedup ratio of the original method declines rapidly, whereas our method exhibits a more gradual decline.}
    \begin{tabular}{ccccccc}
    \hline
        \multirow{2}{*}{\textbf{Model}} & \multirow{2}{*}{\textbf{Batch Size}} & \textbf{Greedy Decoding} & \multicolumn{2}{c}{\textbf{Vanilla Method}}  & \multicolumn{2}{c}{\textbf{Our Method}} \\ 
         &  & TPS & TPS & Speed up & TPS & Speed up \\ 
        \hline
        \multirow{6}{*}{Opt-2.7b} & 1 & 128.31  & 389.61  & 3.04  & ~ & ~ \\
        ~ & 2 & 205.95  & 484.31  & $\underline{2.35}$  & 550.62  & \textbf{2.67}  \\
        ~ & 4 & 362.88  & 545.67  & $\underline{1.50}$  & 707.25  & \textbf{1.95}  \\
        ~ & 8 & 643.38  & 527.27  & $\underline{0.82}$  & 885.83  & \textbf{1.38}  \\
        ~ & 12 & 881.70  & 521.42  & $\underline{0.59}$  & 973.53  & \textbf{1.10}  \\
        ~ & 16 & 1087.33  & 521.16  & $\underline{0.48}$  & 1072.35  & 0.99  \\
        \hline
        \multirow{8}{*}{Opt-13b} & 1 & 59.51  & 219.82  & 3.69  & ~ & ~ \\
        ~ & 2 & 95.33  & 295.37  & $\underline{3.10}$  & 325.79  & \textbf{3.42}  \\
        ~ & 4 & 160.31  & 374.24  & $\underline{2.34}$  & 441.02  & \textbf{2.75}  \\
        ~ & 8 & 278.70  & 379.05  & $\underline{1.36}$  & 578.22  & \textbf{2.07}  \\
        ~ & 12 & 375.74  & 371.32  & $\underline{0.99}$  & 610.32  & \textbf{1.62}  \\
        ~ & 16 & 468.58  & 378.50  & $\underline{0.81}$  & 649.84  & \textbf{1.39}  \\
        ~ & 20 & 540.07  & 390.93  & $\underline{0.72}$  & 742.46  & \textbf{1.37}  \\
        ~ & 24 & 593.53  & 401.34  & $\underline{0.68}$  & 779.77  & \textbf{1.31}  \\
        \hline
    \end{tabular}
    \label{tab:llma_diff_models}
\end{table*}

\subsection{Implementation details}

\textbf{Base Speculative Decoding Methods.}
The efficacy of our approach is evaluated through two fundamental speculative decoding methods.
These include LLMA~\cite{yang2023inference}, a retrieval-based method, and the draft model prediction method~\cite{leviathan2023fast}, which employs draft models to predict.
In the LLMA method, the match length is set to 2 and the copy length to 7.
In the draft model prediction method, the draft model is employed to predict 4 tokens.

\noindent \textbf{Models and Datasets.}
We adopt the Opt~\cite{zhang2022opt} Series models, including Opt-2.7b, Opt-6.7b, and Opt-13b.~\footnote{As the FasterTransformer framework itself does not support the Llama model, we did not utilize the more popular model like Llama for testing purposes.}
For the draft model prediction method, we utilized Opt-125m as the draft model.
The test data set comprised a total of 480 pieces of data selected from the CNN/Daily Mail Test subset~\cite{see-etal-2017-get}.
In our experiments, we utilize a single A100 GPU for the Opt-2.7b and 6.7b models, while employing two GPUs for the Opt-13b model.
All experimental results were subjected to three independent tests and the mean values were calculated.
We also conducted experiments on GSM8K~\cite{cobbe2021training} and MT-bench~\cite{zheng2023judging} datasets, the details of which can be found in Appendix~\ref{sec:appendix_experiments_on_other_datasets}.

\noindent \textbf{Metrics.} 
In order to ascertain the speed of a given method, we employ the tokens per second as an indicator. 
Furthermore, the speed up ratio represents the multiple between the use of the speculative decoding method and its absence.
Given that the generation length of the CNN/Daily Mail Dataset is relatively brief (less than 128), we limit our consideration to the incremental decoding process.
In speculative decoding, the average acceptance length is a significant metric, with a larger average acceptance length often indicative of a higher speedup ratio.
Since some padding tokens need to be added in the vanilla multi-sample speculative decoding method, the average padding ratio is also a significant metric. 

\noindent \textbf{Specific code implementation.}
Our proposed method in question necessitates the alteration of the CUDA kernel. 
And we implemented our method on the FasterTransformer~\cite{NVIDIA2021FasterTransformer} framework, which is a widely used C++ acceleration library that facilitates the implementation of our method.
The proposed methods were implemented by modifying the Python calling interface and the CUDA kernels.
Further details are available in the open-source repository.

\subsection{Experiments using LLMA}

In this section, LLMA is adopted as the basic method of speculative decoding.

\noindent \textbf{Ablation study on two Key components: unpad KV cache and unpad input tokens.}
As illustrated in Table~\ref{tab:llma_ablation},
the opt-6.7b model was employed to conduct ablation experiments on two key methods.
Firstly, it can be observed that under varying batch sizes, our method exhibits a superior speedup compared to the vanilla method. When the batch size was set to 8, our method achieved a speedup of 2.17 times, whereas the vanilla method only achieved a speedup of 1.37 times.
Secondly, both sub-methods are of significance, with "unpad KV cache" playing a particularly pivotal role.

\begin{figure}[!h]
    \centering
	\includegraphics[width=0.85\linewidth]{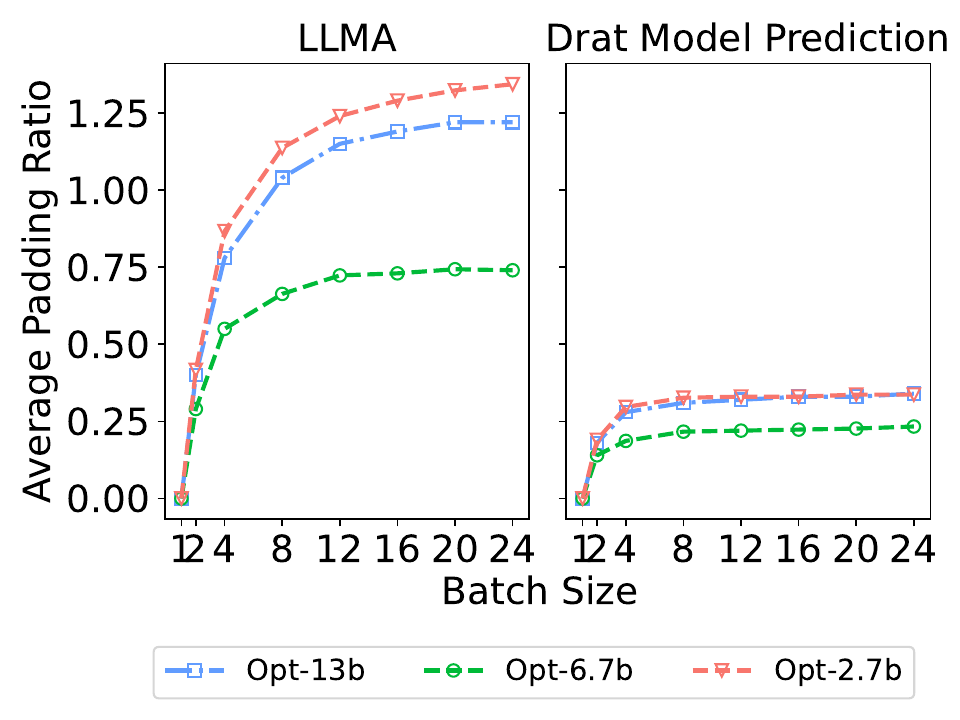}
    \caption{The average padding ratio utilizing the vanilla multi-sample method. 
    The average padding ratio represents the amount of redundant computation, and an increase in this ratio will result in a proportional reduction in speedup.
    }
    \label{img:fig4_batched_vanilla_padding_ratio}
    \vspace{-15pt}
\end{figure}

\noindent \textbf{Experiments on different model sizes.}
We conducted experiments on two models of other sizes, namely opt-2.7b and opt-13b. 
As illustrated in Table~\ref{tab:llma_diff_models}, 
the two smaller models exhibit higher speedup ratios when utilising our method in comparison to the vanilla method, regardless of the varying batch sizes.
With a batch size of 12, the opt-13b model achieved a 1.62x speedup, whereas the vanilla method exhibited no acceleration and was outperformed by the greedy decoding method.
As Figure~\ref{img:fig4_batched_vanilla_padding_ratio} shows, 
the average padding ratio here exceeds 115\%,
highlighting the primary reason for the vanilla method's ineffectiveness.

\begin{table*}[ht]
    \centering
    \caption{Evaluating the effectiveness of our method on three models of different sizes using draft model prediction, with opt-125m model as the draft model.
    In models of varying sizes, our method exhibits a greater speedup than the vanilla method.}
    \begin{tabular}{ccccccc}
    \hline
        \multirow{2}{*}{\textbf{Model}} & \multirow{2}{*}{\textbf{Batch Size}} & \textbf{Greedy Decoding} & \multicolumn{2}{c}{\textbf{Vanilla Method}}  & \multicolumn{2}{c}{\textbf{Our Method}} \\ 
        ~ & ~ & TPS & TPS & Speed up & TPS & Speed up \\ 
        \hline
        \multirow{5}{*}{Opt-2.7b} & 1 & 128.31  & 213.08  & 1.66  & ~ & ~ \\
        ~ & 2 & 205.95  & 306.42  & $\underline{1.49}$  & 318.34  & \textbf{1.55}  \\
        ~ & 4 & 362.88  & 452.65  & $\underline{1.25}$  & 502.58  & \textbf{1.38}  \\
        ~ & 8 & 643.38  & 611.80  & $\underline{0.95}$  & 777.10  & \textbf{1.21}  \\
        ~ & 12 & 881.70  & 685.64  & $\underline{0.78}$  & 944.82  & \textbf{1.07}  \\
        \hline
        \multirow{8}{*}{Opt-6.7b} & 1 & 79.44  & 177.69  & 2.24  & ~ & ~ \\
        ~ & 2 & 137.49  & 273.53  & $\underline{1.99}$  & 285.82  & \textbf{2.08}  \\
        ~ & 4 & 257.37  & 432.38  & $\underline{1.68}$  & 485.98  & \textbf{1.89}  \\
        ~ & 8 & 468.89  & 582.28  & $\underline{1.24}$  & 782.75  & \textbf{1.67}  \\
        ~ & 12 & 644.47  & 635.49  & $\underline{0.99}$  & 945.47  & \textbf{1.47}  \\
        ~ & 16 & 774.59  & 675.88  & $\underline{0.87}$  & 1063.31  & \textbf{1.37}  \\
        ~ & 20 & 863.32  & 709.52  & $\underline{0.82}$  & 1161.78  & \textbf{1.35}  \\
        ~ & 24 & 936.45  & 728.04  & $\underline{0.78}$  & 1258.50  & \textbf{1.34} \\
        \hline
        \multirow{8}{*}{Opt-13b} & 1 & 59.51  & 137.62  & 2.31  & ~ & ~ \\
        ~ & 2 & 95.33  & 201.51  & $\underline{2.11}$  & 223.16  & \textbf{2.34}  \\
        ~ & 4 & 160.13  & 305.09  & $\underline{1.91}$  & 359.14  & \textbf{2.24}  \\
        ~ & 8 & 278.70  & 452.15  & $\underline{1.62}$  & 576.97  & \textbf{2.07}  \\
        ~ & 12 & 375.74  & 494.12  & $\underline{1.32}$  & 728.86  & \textbf{1.94}  \\
        ~ & 16 & 468.58  & 532.73  & $\underline{1.14}$  & 846.73  & \textbf{1.81}  \\
        ~ & 20 & 540.07  & 578.33  & $\underline{1.07}$  & 913.09  & \textbf{1.69}  \\
        ~ & 24 & 593.53  & 612.64  & $\underline{1.03}$  & 974.63  & \textbf{1.64} \\
        \hline
    \end{tabular}
    \label{tab:sd_diff_models}
\end{table*}

\begin{figure}[!h]
    \includegraphics[width=0.9\linewidth]{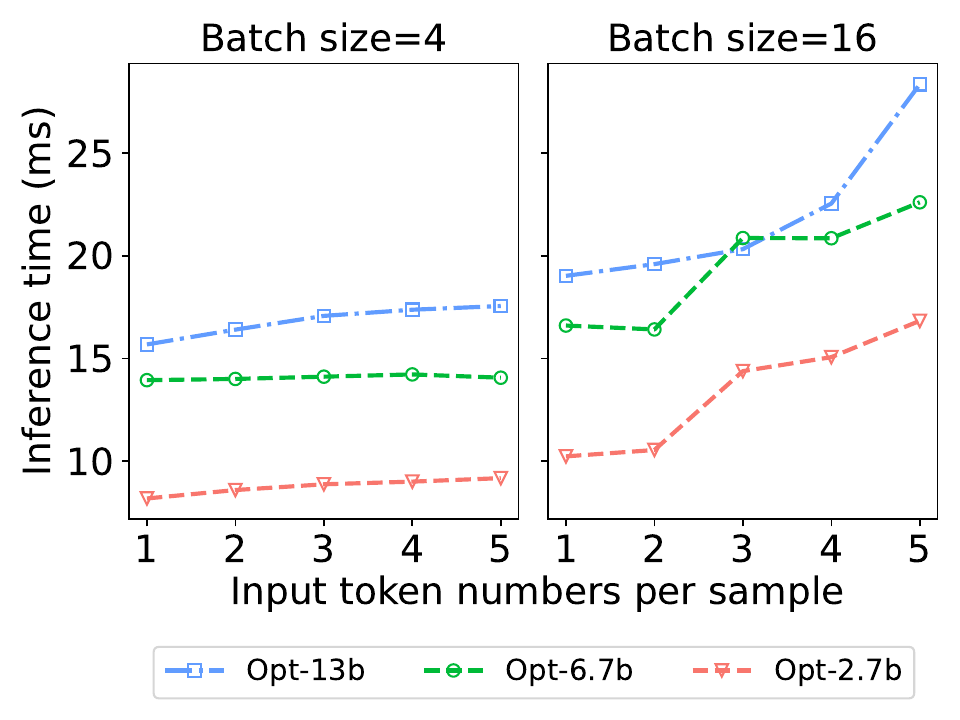}
    \caption{
    The inference time of different numbers of input tokens per sample under different batch sizes. 
    We set the number of existing tokens in each sample to 512.
    When the number of input tokens per sample is varied with a batch size of 4, the inference time remains essentially unchanged. 
    However, when the batch size is increased to 16, the inference time changes significantly.
    }
    \label{fig:input_tokens_speed}
    \vspace{-15pt}
\end{figure}

\subsection{Experiments using Draft Model Prediction}

In this section, the draft model prediction method is adopted as the basic method of speculative decoding.
It is important to note that when utilizing the draft model prediction approach, the number of predictions for each sample is identical. Consequently, only "unpad KV cache" are employed in this section.

As illustrated in Table~\ref{tab:sd_diff_models}, 
we utilize the opt-125m model as the draft model, and test three models of varying sizes.
Our method exhibits a superior speedup compared to the vanilla method across diverse models and varying batch sizes.
As illustrated in Figure~\ref{img:fig4_batched_vanilla_padding_ratio}, 
The opt-6.7b model, with batch size set to 8, exhibits a significant increase in the number of padding tokens, exceeding 60\% using LLMA and exceeding 20\% using draft model prediction.
This explains why vanilla method have serious speedup degradation in multi-sample cases.

\begin{figure*}[ht]
    \centering
    \subfigure[Probability density]{
	    \includegraphics[width=0.23\linewidth]{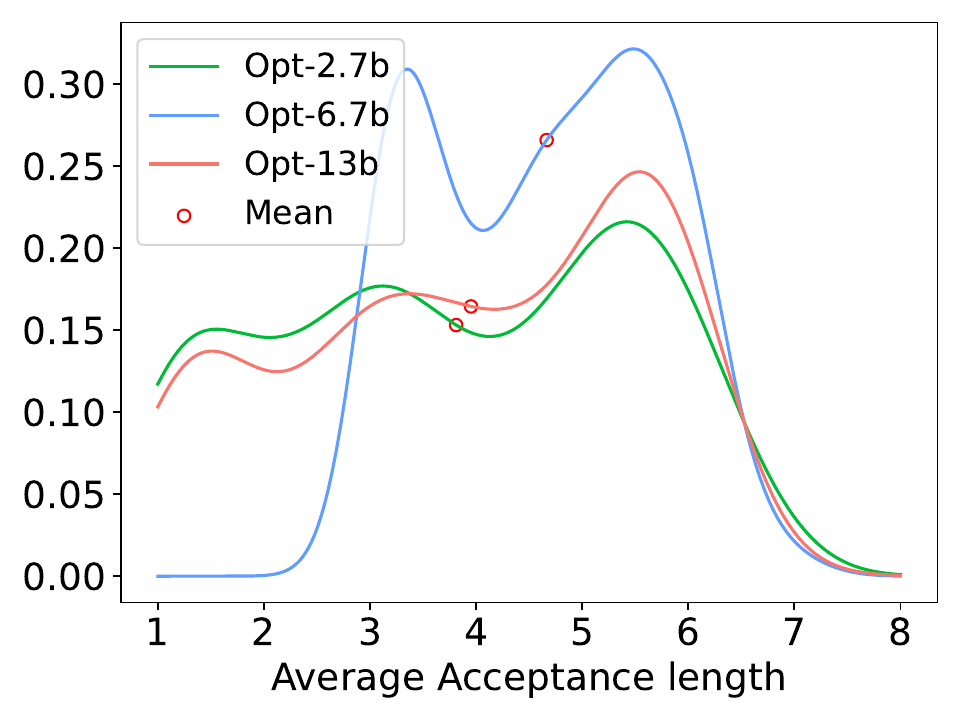}
	    \label{img:fig4_batched_accept_length_pading_length_llma_compression_hist}
	}
	\subfigure[Minimum average acceptance length]{
	    \includegraphics[width=0.23\linewidth]{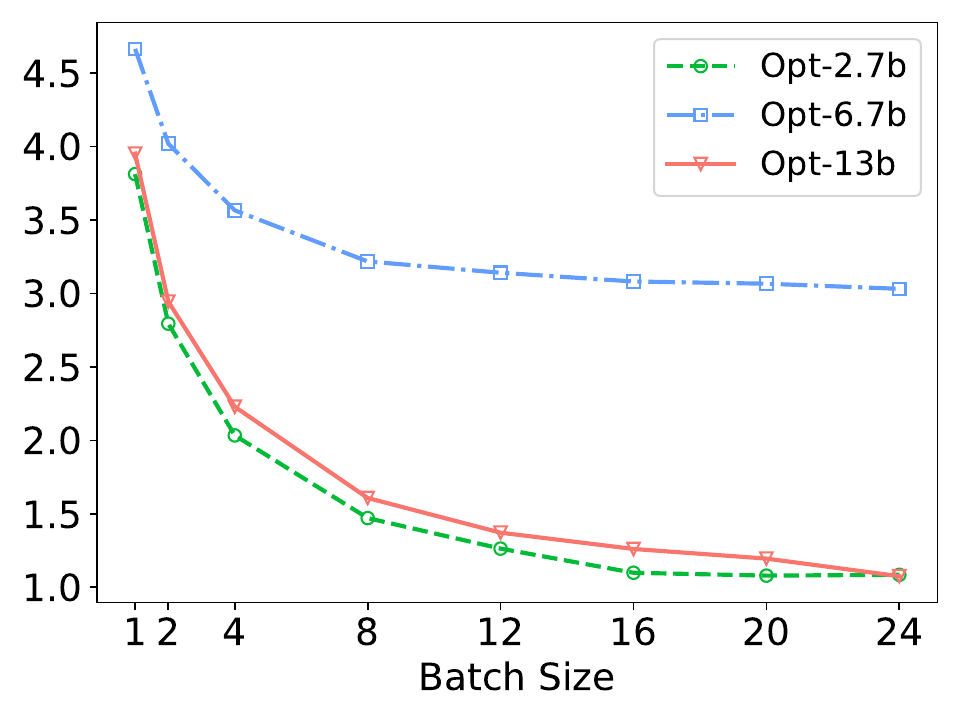}
	    \label{img:fig4_batched_accept_length_pading_length_llma_compression}
	}
	\subfigure[Probability density]{
	    \includegraphics[width=0.23\linewidth]{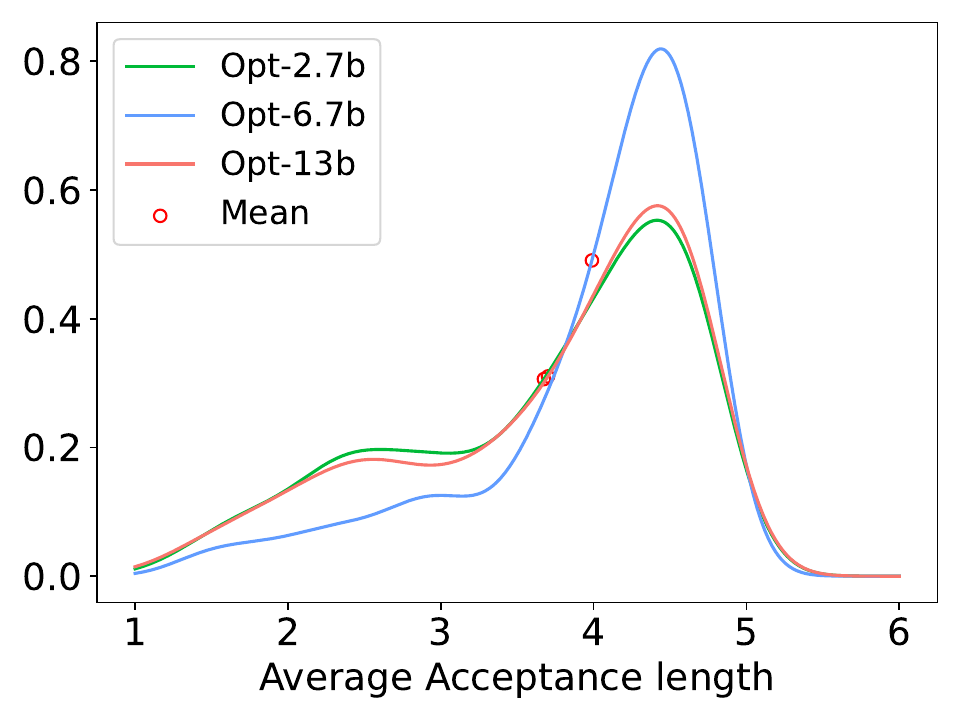}
	    \label{img:fig4_batched_accept_length_pading_length_sd_compression_hist}
	}
	\subfigure[Minimum average acceptance length]{
	    \includegraphics[width=0.23\linewidth]{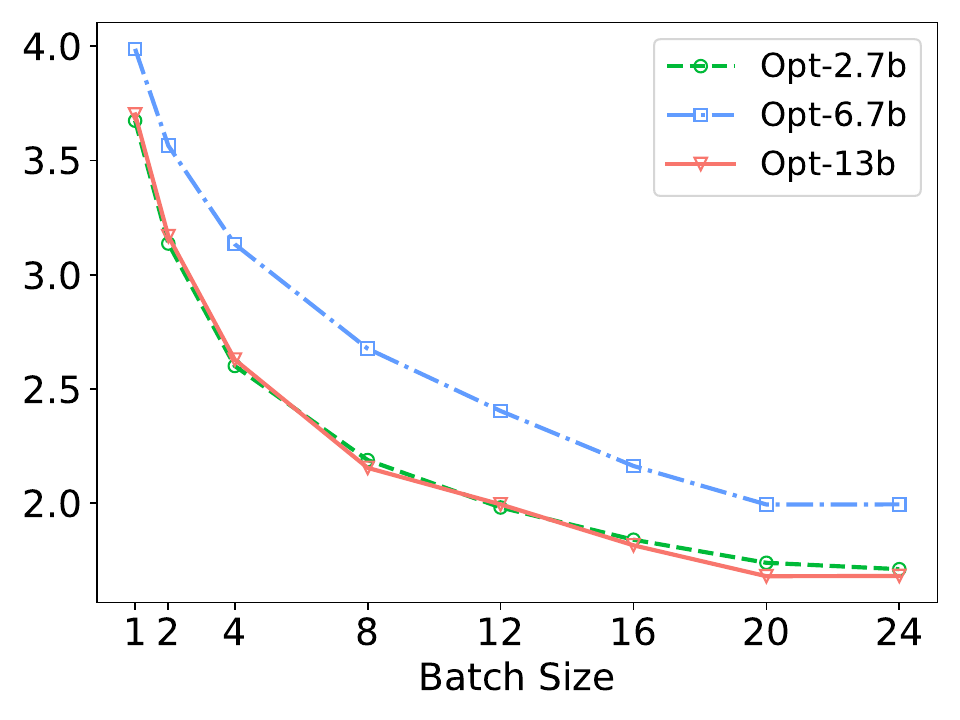}
	    \label{img:fig4_batched_accept_length_pading_length_sd_compression}
	}
    \caption{ The average acceptance length in sigle-sample and multi-sample scenarios.
    Figure (a)(b) employ LLMA as the basic speculative decoding method, while Figure (c)(d) utilize the draft model prediction method, utilising opt-125m as the draft model.
    Figure (a)(c) illustrates the probability density function of the average acceptance length of distinct samples, with batch size set to 1. It is evident that the average acceptance length of different samples exhibits considerable variability. 
    Figure (b)(d) illustrates the reduction in the minimum average acceptance length within a batch.
    Given that the average acceptance length of different samples within a batch are disparate, 
    the minimum value better represents the overall batch's acceleration effect.
    }
    \label{img:fig4_batched_accept_length_pading_length}
\end{figure*}

\subsection{Analysis of Speedup Decrease with Multi-sample}

As illustrated in Table~\ref{tab:llma_ablation}, it is evident that the speedup ratio exhibits a decline in the context of multiple samples. When the batch size is set to 4, the speedup ratio is 2.83, while when the batch size is set to 16, the speedup ratio is 1.63. 
Similar conclusions can be drawn from Table~\ref{tab:sd_diff_models}.

We have identified two factors contributing to the reduction in the speedup ratio. 
Firstly, when the batch size is sufficiently large and multiple tokens are processed simultaneously, the individual inference time for LLMs escalates considerably.
As illustrated in Figure~\ref{fig:input_tokens_speed}, the inference time for the opt-6.7b model with a batch size of 16 is 22.6 milliseconds for the processing of five tokens per sample, whereas for a single token, it is 16.6 milliseconds, which is 1.36 times slower.

Secondly, the principal reason for this decline in performance is the considerable disparity in the speedup across different samples.
The average acceptance length is positively correlated with the speedup ratio.
As illustrated in Figure~\ref{img:fig4_batched_accept_length_pading_length_llma_compression_hist},
the average acceptance length difference of opt-2.7b/13b on different samples is greater than that of opt-6.7b when the LLMA method is employed. 
Figure~\ref{img:fig4_batched_accept_length_pading_length_sd_compression_hist} shows that the discrepancies in the average acceptance length across models were relatively modest when utilising a draft model to predict.

As batch size increases, the minimum average acceptance length within the batch decreases.
As illustrated in Figure~\ref{img:fig4_batched_accept_length_pading_length},
a comparison of the opt-6.7b and opt-2.7b models reveals that the speedup ratio of the latter is more uneven on the test samples.
As the batch size increases, the minimum average acceptance length within the batch decreases at a faster rate, although their speedup ratios are similar when the batch size is equal to one.

In order to maintain the speedup ratio in the multiple samples cases, the most straightforward method is to ensure that the speedup ratios of different samples are similar under the basic speculative decoding method.
However, the optimal solution to this issue is dynamic batching~\cite{yu2022orca}, which entails replacing a finished sample in the batch with a new sample once it has been completed, rather than waiting for all samples in the batch to be completed before proceeding to the next inference.
The implementation of dynamic batching is expected to enhance the efficiency of multi-sample processing, with the potential for achieving comparable speedup to that in single-sample cases.

\section{Conclusions}

In this paper, we present the first study of multi-sample speculative decoding.
we introduce an effective method, called EMS-SD.
EMS-SD is an effective solution to the inconsistency problem of different samples in the prediction and verification stages, without the need of padding tokens.
The proposed method is flexibly integrated with almost any basic speculative decoding method. 
Extensive comparisons show that EMS-SD exhibits superior performance compared to the vanilla method in multi-sample speculative decoding.

\section*{Limitations}

This work has four limitations:
1) Theoretical evidence indicates that dynamic batching may serve to mitigate the performance degradation that occurs in multi-sample speculative decoding.
However, this has not been empirically validated.
Subsequent experiments will assess the efficacy of multi-sample speculative decoding in conjunction with dynamic batching.
2) The potential negative impact of non-contiguous memory accesses on performance was not considered. 
In batched greedy decoding, the memory access between different samples is continuous.
However, in the proposed method, due to the varying lengths of different samples, the memory access is not continuous.
This may have a negative effect on acceleration.
3) Although our method is independent of the inference framework, we have not yet implemented our method on frameworks such as PyTorch~\cite{paszke2019pytorch} or vLLM~\cite{kwon2023efficient}.
This undoubtedly limits the ease of use of our method.
In subsequent work, we will consider implementing our method in these frameworks.
4) Tree decoding will further accelerate speculative decoding, which has been widely verified in the single-sample speculative decoding~\cite{miao2023specinfer, cai2024medusa, liu2024kangaroo, li2024eagle}.
Nevertheless, the efficacy of integrating tree decoding with multi-sample speculative reasoning has yet to be validated.
Future experiments will evaluate the effectiveness of the multi-sample speculative decoding when integrated with tree decoding.

\bibliography{refs}
\bibliographystyle{acl_natbib}

\appendix


\section{Preliminaries}

\begin{figure*}[ht]
    \centering
	\subfigure[Maximum Acceptance Length]{
	    \includegraphics[width=0.31\linewidth]{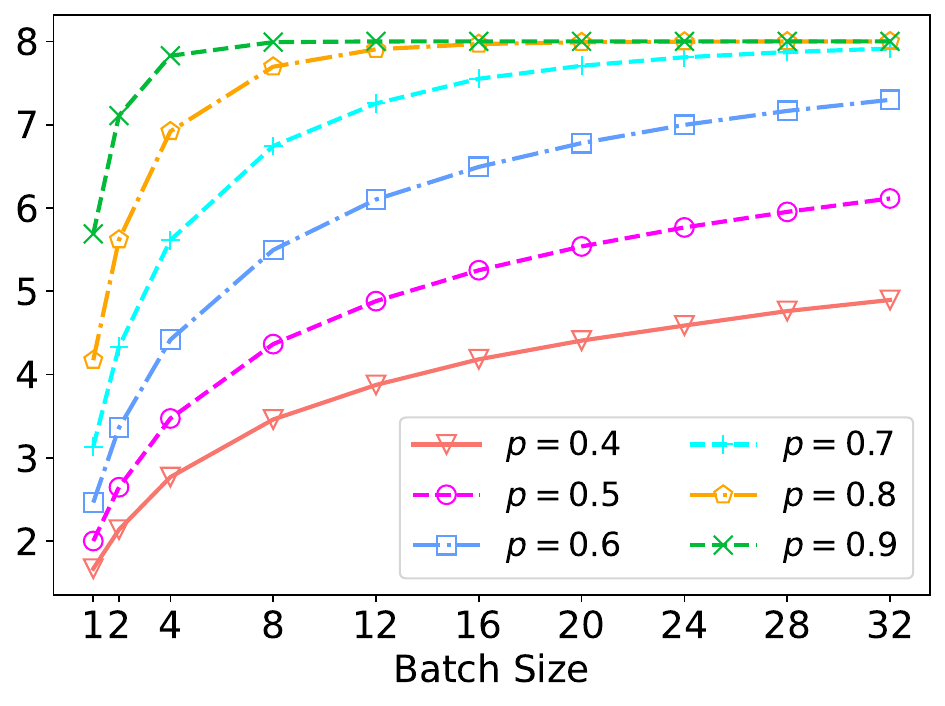}
	}
	\subfigure[Average Padding Length]{
	    \includegraphics[width=0.31\linewidth]{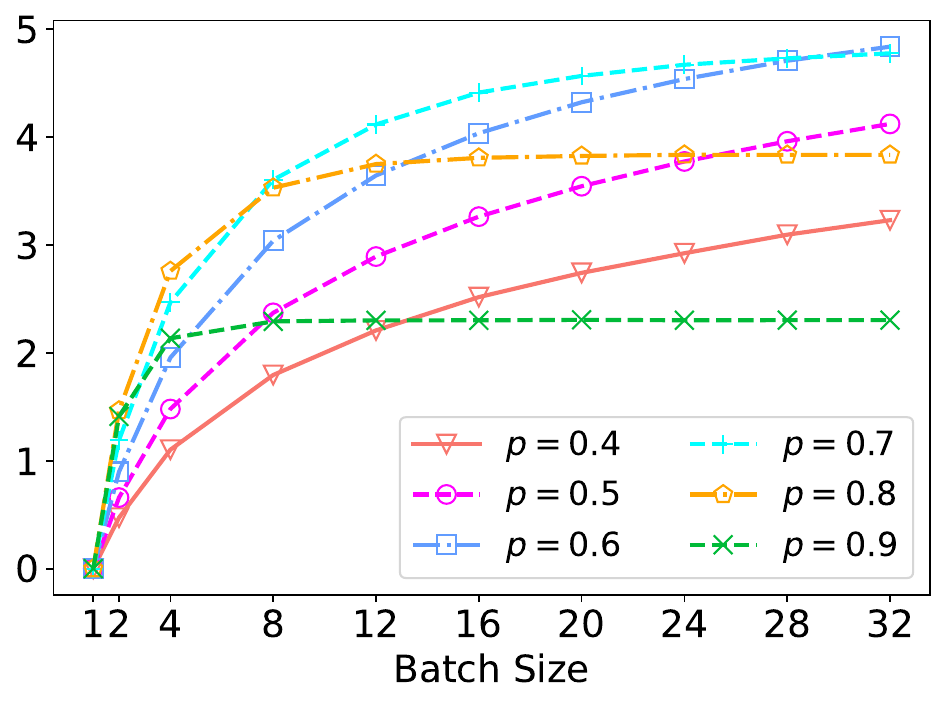}
	}
	\subfigure[Average Padding Ratio]{
	    \includegraphics[width=0.31\linewidth]{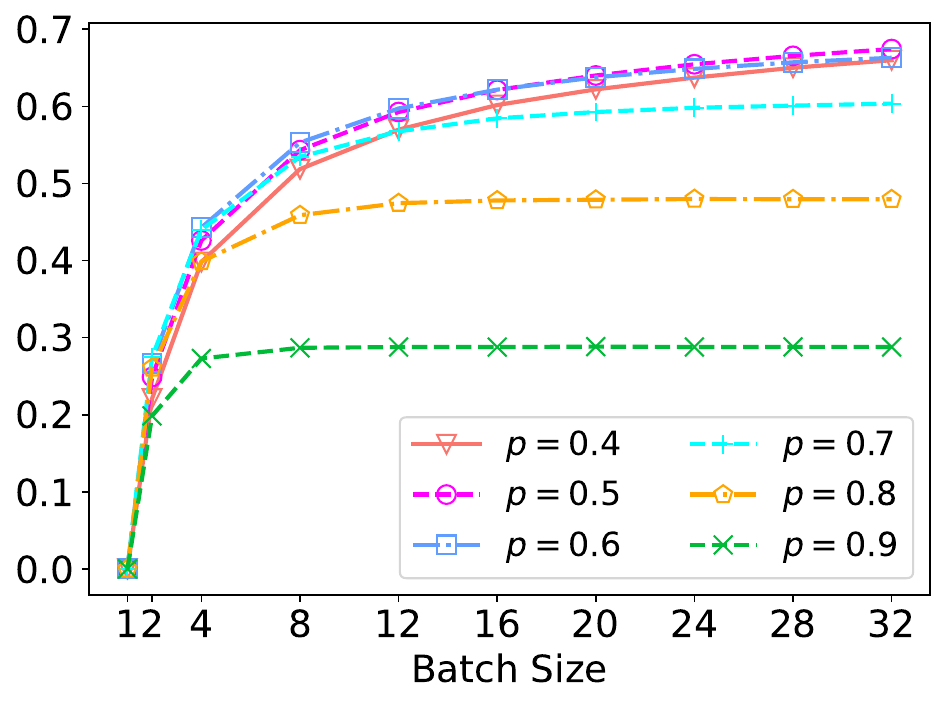}
	}
    \caption{Numerical simulation of the expected value of three variables: the maximum acceptance length $\tau_{max}$, the average padding length $\overline{\delta}$ and the average padding ratio $\overline{r}$. }
    \label{img:fig2_vanilla_accept_length_pading_length}
\end{figure*}

\textbf{Autoregressive Decoding.} Autoregressive large language models (LLMs) $P$ generates a token at each step. 
Let $x$ be the sequence of tokens, with $x_j$ denoting the token at position $j$.
The probability distribution of the the token at position $i$ over vocabulary $V$, $y_i$, is contingent upon the preceding tokens.
Consequently, $y_i$ can be expressed as Equation~\ref{eq:auto_LLM}.

\begin{equation}
    y_i \sim P(y|x_{[0,i)})
    \label{eq:auto_LLM}
\end{equation}

For greedy decoding, the subsequent token is selected according to the maximum value of the probability distribution.

\begin{equation}
    x_i = \mathop{\arg\max}\limits_{y \in V } y_i
    \label{eq:auto_LLM_greedy}
\end{equation}

\noindent \textbf{Single-sample Speculative Decoding.}
With regard to Speculative Decoding, the process can be divided into two stages: prediction and verification.
In the prediction stage, a prediction method, $f$, is employed to predict the subsequent $k$ tokens $d_i,..,d_{i+k-1}$ at each step.

\begin{equation}
    d_i,..,d_{i+k-1} = f (x_{[0,i)} )
    \label{eq:auto_LLM_sd_prediction}
\end{equation}

In the verification stage, these $k$ predicted tokens are simultaneously input to the LLMs, together with existing tokens. This enables the LLMs to generate $k+1$ tokens in a single decoding.


\begin{align}
  x_j = & \mathop{\arg\max}\limits_{y \in V } P ( y|x_{[0,i)}, d_{[i,j)} ),  \notag \\ 
     &  i \le j < i+k+1 
  \label{eq:auto_LLM_sd_verification}
\end{align}

The output token $x_i$ is identical to the result generated by the Autoregressive method.
Nevertheless, it is essential to ascertain the remaining $k$ tokens ($x_j, i+1 \le j < i+k+1$) to ascertain their acceptability.
The acceptance length $\tau$ can be calculated using the Equation~\ref{eq:auto_LLM_sd_verification_accept}.
Since $d_{i+k}$ is undefined, it follows that $x_{i+k}$ and $d_{i+k}$ are always unequal.

\begin{align}
    \tau = & \mathop{\arg\max}\limits_{j} \{x_{i+j-1} \ne d_{i+j-1} | \notag \\ 
       & x_{i+m-1} = d_{i+m-1}, 1 \le m < j \}, \notag \\
        &  1 \le j \le k+1
    \label{eq:auto_LLM_sd_verification_accept}
\end{align}

Consequently, the LLMs is capable of accepting $\tau$ tokens simultaneously, rather than just one, within a similar timeframe.
It is important to note that the average acceptance length $\overline{\tau}$ and the speedup ratio are closely related.
As the average acceptance length increases, the speedup ratio also rises.

\section{Theoretical Analysis of the Impact of Padding Tokens in Vanilla Multi-sample Speculative Decoding}
\label{sec:appendix_analysis_vanilla_padding_tokens}

As mentioned in Section~\ref{sec:vanilla_method}, the introduction of additional padding tokens in the vanilla method will result in a reduction in speedup ratio in multi-sample cases.
We assume that $k$ tokens in the prediction stage, and the prediction accuracy of the next token is $p$, thus the accepted length $\tau$ conforms to the geometric distribution.
The probability mass function of $\tau$ can be formulated as Equation~\ref{eq:vanilla_sd_probability}. 
And the expected value $E(\tau)$ is formulated as Equation~\ref{eq:vanilla_sd_probability_e_var}. 

\begin{equation}
    P(\tau = k) = p^{k-1}(1-p), k=1,2,3,4,...
    \label{eq:vanilla_sd_probability}
\end{equation}
\begin{equation}
    E(\tau) = \frac{1}{1-p} 
    \label{eq:vanilla_sd_probability_e_var}
\end{equation}

\begin{table*}[ht]
    \centering
    \caption{Performance of vanilla method v.s. our EMS-SD method using GSM8K~\cite{cobbe2021training} and MT-Bench~\cite{zheng2023judging} dataset.}
    \begin{tabular}{ccccccc}
    \hline
        \multirow{2}{*}{\textbf{Datasets}} & \textbf{Batch} & \textbf{Greedy Decoding} & \multicolumn{2}{c}{\textbf{Vanilla Method}}  & \multicolumn{2}{c}{\textbf{Our Method}} \\ 
        ~ & \textbf{Size} & TPS & TPS & Speed up & TPS & Speed up \\
        \hline
        \multirow{8}{*}{GSM8K} & 1 & 79.11  & 186.64  & 2.36  & ~ & ~ \\
        ~ & 2 & 89.24  & 99.96  & $\underline{1.12}$  & 156.77  & \textbf{1.76}  \\
        ~ & 4 & 163.54  & 183.14  & $\underline{1.12}$  & 273.70  & \textbf{1.67}  \\
        ~ & 8 & 310.15  & 318.07  & $\underline{1.03}$  & 467.35  & \textbf{1.51}  \\
        ~ & 12 & 436.10  & 389.36  & $\underline{0.89}$  & 593.25  & \textbf{1.36}  \\
        ~ & 16 & 554.62  & 413.12  & $\underline{0.74}$  & 631.31  & \textbf{1.14}  \\
        ~ & 20 & 646.97  & 470.98  & $\underline{0.73}$  & 720.56  & \textbf{1.12}  \\
        ~ & 24 & 722.49  & 514.06  & $\underline{0.71}$  & 798.40  & \textbf{1.11}  \\
        \hline
        \multirow{8}{*}{MT-Bench} & 1 & 59.59  & 151.72  & 2.55  & ~ & ~ \\
        ~ & 2 & 112.83  & 257.65  & $\underline{2.28}$  & 300.03  & \textbf{2.66}  \\
        ~ & 4 & 202.57  & 442.06  & $\underline{2.18}$  & 531.44  & \textbf{2.62}  \\
        ~ & 8 & 374.14  & 642.18  & $\underline{1.72}$  & 931.24  & \textbf{2.45}  \\
        ~ & 12 & 519.41  & 699.02  & $\underline{1.35}$  & 1175.50  & \textbf{2.26}  \\
        ~ & 16 & 654.17  & 758.04  & $\underline{1.16}$  & 1310.58  & \textbf{2.00}  \\
        ~ & 20 & 777.47  & 881.30  & $\underline{1.13}$  & 1494.98  & \textbf{1.92}  \\
        ~ & 24 & 856.29  & 848.97  & $\underline{0.99}$  & 1603.13  & \textbf{1.87} \\
        \hline
    \end{tabular}
    \label{tab:sd_diff_datasets}
\end{table*}

\begin{table*}[h]
    \centering
    \caption{Performance of vanilla method v.s. our EMS-SD method using three different datasets.}
    \begin{tabular}{cccccc}
    \hline
        \multirow{2}{*}{\textbf{Datasets}} & \textbf{Average} & \textbf{Max} & \multicolumn{2}{c}{\textbf{Vanilla Method}} & \textbf{Our Method} \\ 
        ~ & \textbf{Input Length} & \textbf{Output Length} & \textbf{Padding Ratio} & \textbf{Speed Up} & \textbf{Speed Up} \\
        \hline
        CNN/Daily Mail  & 744.94    & 128   & 30.82  & $\underline{1.62}$  & \textbf{2.07} \\
        GSM8K           & 1240.99   & 128   & 89.12  & $\underline{1.03}$  & \textbf{1.51} \\
        MT-Bench        & 220.84    & 1024  & 13.23  & $\underline{1.72}$  & \textbf{2.49} \\
        \hline
    \end{tabular}
    \label{tab:sd_diff_datasets_compare}
\end{table*}

If $b$ samples are inferred simultaneously (batch size is set to $b$), the maximum acceptance length $\tau_{max}$ and the average padding length $\overline{\delta}$ can be expressed as Equation~\ref{eq:vanilla_sd_probability_bs_tau} and~\ref{eq:vanilla_sd_probability_bs_delta}.
Furthermore, we define the average padding ratio $\overline{r}$, which is the ratio of $\overline{\delta}$ and $\tau_{max}$, as shown in Formula~\ref{eq:vanilla_sd_probability_bs_ratio}.
It can be demonstrated that as the value of $\overline{r}$ increases, the proportion of padding also increases, resulting in a greater waste of computational and memory access overhead. 
It can be observed that as the average padding ratio increases, the negative impact on acceleration also increases.

\begin{equation}
    \tau_{max} = max(\tau_0, \tau_1,..., \tau_{b-1})
    \label{eq:vanilla_sd_probability_bs_tau}
\end{equation}
\begin{equation}
    \overline{\delta} = \tau_{max} - \frac{1}{b}(\tau_0 + \tau_1 + ... + \tau_{b-1})
    \label{eq:vanilla_sd_probability_bs_delta}
\end{equation}
\begin{equation}
    \overline{r} = \overline{\delta} / \tau_{max}
    \label{eq:vanilla_sd_probability_bs_ratio}
\end{equation}

The probability mass function of $\tau_{max}$ can be expressed as Equation~\ref{eq:vanilla_sd_probability_bs_tau_max}. 

\begin{align}
     &P( \tau_{max}=k) = \notag \\ 
        & \begin{cases}
        (1-p^k)^b - (1-p^{k-1})^b, & \text{if } k > 1 \\
        (1-p)^b, & \text{if } k = 1
        \end{cases}
    \label{eq:vanilla_sd_probability_bs_tau_max}
\end{align}

The expected value of $\tau_{max}$ and $\overline{\delta}$ are challenging to express in a concise manner.
This is why we employed numerical simulation method.
As shown in Figure~\ref{img:fig2_vanilla_accept_length_pading_length}, 
We show the expected value of $\tau_{max}$, $\overline{\delta}$ and $\overline{r}$ as they vary with the prediction accuracy of next token $p$ and batch size $b$.
In practical applications, the maximum acceptance length is constrained by the limitations of the predicted length.
In this figure, we limited the maximum acceptance length to 8.

Two conclusions can be drawn from Figure~\ref{img:fig2_vanilla_accept_length_pading_length}.
Firstly, the maximum acceptance length and the average padding length both increase as the prediction accuracy of next token and the batch size increase.
Secondly, when the maximum acceptance length is limited, the higher the prediction accuracy of next token, the lower the average padding ratio.
In particular, when the prediction accuracy of next token is below 0.8 and the batch size is greater than 8, the padding ratio increases rapidly to exceed 50\%.
However, even if the prediction accuracy reaches 90\%, 30\% of the computational and memory access overhead are still wasted.

\section{Experiments on Other Datasets}
\label{sec:appendix_experiments_on_other_datasets}

As demonstrated in Rebuttal Table~\ref{tab:sd_diff_datasets}, the efficacy of our multi-sample speculative decoding method was evaluated on the GSM8K~\cite{cobbe2021training} and MT-Bench~\cite{zheng2023judging} datasets. 
The basic single-sample speculative inference method employed here is draft model prediction.
In particular, opt-13b was utilized as the large language model, while opt-125m were employed as small models.
It was observed that when the batch size exceeded 8, the benefit of our method exceeded 45

As demonstrated in Table~\ref{tab:sd_diff_datasets_compare}, we evaluate the impact of three distinct datasets on our method when the batch size was set to 8.
It can be observed that our EMS-SD method demonstrates enhanced performance when applied to diverse datasets. 
The magnitude of this improvement is contingent upon the extent of reduction in the computational and memory access overhead.
The increased computational and memory access overhead during multi-sample speculative decoding is influenced by three key factors: input length, output length, and the prediction imbalance between samples, which is reflected in the padding ratio.
It can be demonstrated that the greater the computational and memory access overhead, the more significant the gain of our method.

\section{The Specific Process of Unpad Input Tokens}
\label{sec:appendix_algorihtm_of_unpad_input_tokens}

Before being fed into the Transformer model, all input tokens are merged into a single sequence, and the quantity of input tokens for each sample is documented.This process is detailed in Algorithm~\ref{alg:concatenate_input_tokens}.

Furthermore, when computing the attention output, the original batch index and sequence position of each token are restored in the CUDA kernel.

This process is detailed in Algorithm~\ref{alg:concatenate_input_tokens_restore_origin_bs}.
In addition, modifications will be required to the grid responsible for invoking the CUDA kernel during the attention calculation process.
Equation~\ref{eq:grid_change} illustrateds the specific alterations.
In the context of the CUDA kernel, the value of $blockIdx.y$ represents the index of the current token among all inputs.

\begin{align}
    & grid(num\_heads, batch\_size) \rightarrow \notag \\ 
    & grid (num\_heads, total\_input\_token\_nums)
    \label{eq:grid_change}
\end{align}

\begin{algorithm*}
    \DontPrintSemicolon 
    \caption{Concatenate the input tokens of different samples}
    \label{alg:concatenate_input_tokens}
    \KwData{$list\_of\_input\_tokens$}
    \KwResult{$concatenated\_input\_tokens$, $token\_nums\_per\_sample$, $total\_input\_token\_nums$}
    $batch\_size$ = len( $list\_of\_input\_tokens$ )\;
    $concatenated\_input\_tokens$ = []\;
    $token\_nums\_per\_sample$ = [0 for \_ in range($batch\_size$)]\;
    $total\_input\_token\_nums$ = 0 \;
    \For{$i = 0$ \KwTo $batch\_size-1$}{
        $total\_input\_token\_nums$ += len($list\_of\_input\_tokens[i]$) \;
        $token\_nums\_per\_sample[i]$ = len($list\_of\_input\_tokens[i]$) \;
        $concatenated\_input\_tokens$.extend($list\_of\_input\_tokens[i]$) \;
    }
\end{algorithm*}

\begin{algorithm*}
    \DontPrintSemicolon 
    \caption{Restore the original batch index and sequence position in CUDA kernels}
    \label{alg:concatenate_input_tokens_restore_origin_bs}
    \KwData{$token\_nums\_per\_sample$, $blockIdx$}
    \KwResult{$original\_batch\_index$, $original\_sequence\_position$}
    $batch\_size$ = len( $token\_nums\_per\_sample$ )\;
    $original\_sequence\_position$ = $blockIdx.y$ \;
    $original\_batch\_index$ = 0 \;
    \For{$i = 0$ \KwTo $batch\_size-1$}{
        \eIf{ $original\_sequence\_position \ge token\_nums\_per\_sample[i]$ }{
            $original\_batch\_index$ += 1 \;
            $original\_sequence\_position$ -= $token\_nums\_per\_sample[i]$
        }{
            break\;
        }
    }
\end{algorithm*}

\end{document}